*Article*

# Machine Learning Techniques for Predicting the Short-Term Outcome of Resective Surgery in Lesional-Drug Resistance Epilepsy

Zahra Jourahmad[1], Jafar Mehvari Habibabadi[1], Houshang Moein[2], Reza Basiratnia[3], Ali Rahmani Geranqayeh[4], Saeed Shiry Ghidary[4], and Seyed-Ali Sadegh-Zadeh[4*]

[1] Kashani Comprehensive Epilepsy Center, Kashani Hospital, Isfahan, Iran
[2] Department of Neurosurgery, School of Medicine, Isfahan University of Medical Sciences, Isfahan, Iran
[3] Department of Radiology, School of Medicine, Isfahan University of Medical Sciences, Isfahan, Iran
[4] Department of Computing, School of Digital, Technologies and Arts, Staffordshire University, Stoke-on-Trent, United Kingdom
Orcid ID: First author: Zahra Jourahmad: https://orcid.org/0000-0002-3728-2887
Senior author: Seyed-Ali Sadegh-Zadeh: https://orcid.org/0000-0001-6197-3410

**Abstract:** In this study, we developed and tested machine learning models to predict epilepsy surgical outcome using noninvasive clinical and demographic data from patients. **Methods:** Seven different categorization algorithms were used to analyze the data. The techniques are also evaluated using the Leave-One-Out method. For precise evaluation of the results, the parameters accuracy, precision, recall and, F1-score are calculated. **Results:** Our findings revealed that a machine learning-based presurgical model of patients' clinical features may accurately predict the outcome of epilepsy surgery in patients with drug-resistant lesional epilepsy. The support vector machine (SVM) with the linear kernel yielded 76.1% in terms of accuracy could predict results in 96.7% of temporal lobe epilepsy (TLE) patients and 79.5% of extratemporal lobe epilepsy (ETLE) cases using ten clinical features. **Significance:** To predict the outcome of epilepsy surgery, this study recommends the use of a machine learning strategy based on supervised classification and selection of feature subsets data mining. Progress in the development of machine learning-based prediction models offers optimism for personalised medicine access.

**Key point box**

We developed and tested a machine learning model to predict epilepsy surgical outcome using noninvasive clinical and demographic features.

A set of three factors, existence of aura, focal to bilateral seizures, and epilepsy duration, accounted for 85% of the model's accuracy.

The Support vector machine (SVM) with the linear kernel yielded 76.1% in terms of accuracy could predict results in 96.7% of TLE patients and 79.5% of ETLE cases.

## 1. Introduction

An epileptic seizure (ES) is caused by a sudden abnormal, self-sustained electrical discharge that occurs in the cerebral networks and usually lasts for less than a few minutes. Patients with epilepsy whose seizures do not successfully respond to antiseizure medication therapy are considered to have drug-resistant epilepsy (DRE). This condition is also referred to as intractable, medically refractory, or pharmacoresistant epilepsy.[1]

Research indicates that for focal drug-resistant epilepsy, brain surgery is the gold standard treatment.[2] Epilepsy surgery is the resection or functional manipulation of part/region of the brain with the aim of alleviating seizures, improving the cognitive function and the quality of life. New preoperative techniques offer the opportunity of improved presurgical planning and selection of cases more likely to be rendered seizure-free by current surgical techniques.[3] Seizure freedom rates, on the other hand, range from 30% to 80% in different studies.[4] Despite of numerous attempts to predict the outcome of epilepsy surgery, there is no reliable attributes of successful surgery outcome. Based on a broad anatomical parcellation a higher success rates ranging from 65% to 73% accounts for seizures originating from temporal lobe (TLE) compared to those arising from the extratemporal regions (ETLE) with success rate of 27% to 60% [33]. The success criteria are determined by inherent and extrinsic elements that function as "positive" and "negative" influences on the outcome. Clinicians have struggled to predict outcome, as interpreting presurgical test results in the analysis of patients with intractable epilepsy is subjective.[5] The lack of a standardised, objective, and effective tool for evaluating and predicting surgical success, makes pre-surgery decision-making more difficult. Given that half of the cases failed in surgery, there is a need for accurate outcome prediction tools to assist the patient with preoperative counselling, which reduces the need for referral to a higher care facility by early identification of difficult patient needs.

Some research has questioned whether clinical data, imaging techniques (MRI, PET, SPECT), electroencephalographic function tests (EEG, video-EEG), neuropsychological tests, or a combination of these methods may accurately predict epilepsy surgery. Since epilepsy is a complex condition with many variables, the predictive usefulness of individual variables is limited.[6] On the other hand, machine learning approaches can create models that combine patient unique data to predict the outcome of surgical treatment and thus aid in clinical decision making. [30,31,32] As a result, academics have sought to develop machine learning algorithms as tools for predicting epilepsy.

Recently, significant progress has been achieved in demonstrating that machine learning methods may predict seizure incidence in epilepsy patients. ML algorithms have been used as prediction models for survival, recurrence, symptom improvement, and adverse events in patients undergoing surgery for epilepsy, brain tumor, spinal lesions, neurovascular disease, movement disorders, traumatic brain injury, and hydrocephalus.[7] Some studies also demonstrated a better performance in ML models compared with established prognostic indices and clinical experts.[8]

In this study computational models were generated using supervised classification data mining methodologies to predict if a patient would recover completely following invasive surgery.

## 2. Materials and Methods

In this study, seven different machine learning classification models were used including decision tree, random forest, multi-layer perceptron, logistic regression, k-nearest neighbour, gradient boosting, and support vector machines with various kernels such as linear, polynomial, RBF, and sigmoid. To evaluate the classification performance of each model, we used two validation techniques, i.e. leave-one-out cross-validation which is exhaustive cross-validation, and k-fold cross-validation which is non-exhaustive cross-validation. In leave-one-out cross-validation, each learning set is created by taking all the samples but one, the test set being the sample left out. Therefore, for n samples, we have n different training sets and n different test sets. In k-fold cross-validation, all the samples are divided into K groups of samples. These groups of samples have equal sizes. The model is learned using k-1 folds and only the remaining fold can be used for the test. Accuracy estimations were also performed for each model.

A dataset is imbalanced if the classification categories are not approximately equally represented. Often real-world data sets are predominantly composed of "normal" examples with only a small percentage of "abnormal" or "interesting" examples. In this regard, several methods were researched and the oversampling method was selected finally.

Oversampling methods duplicate examples in the minority class or synthesize new examples from the examples in the minority class. Some of the more widely used and implemented oversampling methods include: Random Oversampling, Synthetic Minority Oversampling Technique (SMOTE), Borderline-SMOTE, Borderline Oversampling with Support Vector Machine (SVM), Adaptive Synthetic Sampling (ADASYN).[9] In all implementation the default configurations in Python were used.

## 3. Dataset Compilation

We carried a retrospective analysis of data at the Kashani Comprehensive Epilepsy Center of Isfahan University of Medical Sciences in Isfahan, Iran. The Bioethics Committee of Isfahan University of Medical Sciences has approved this study under the verification code of IR.MUI.MED.REC.1399.669.

Records of adult patients (age>16 years) with lesional-drug resistance epilepsy who have undergone resective epilepsy surgery were reviewed. Clinical and imaging data were obtained from patients' documentations. For this investigation, 176 records were prepared. Each record consists of 13 useful features, including 9 categorised, 3 numer-ical, and a label. The results of previous studies have demonstrated different and con-tradicting influence of various variables (clinical, imaging, neuropsychological) on the epilepsy surgery outcome [refs]. Here we selected features based on the review of literature, expert opinion and ease of accessibility.

An expert epileptologist assessed the surgery outcome during six months to five years of follow up visits (depending on the date that patients were enrolled in the study). The outcome of surgery was defined as successful/unsuccessful based on the modi-fied Engel score. This dataset is divided into two different groups: 128 patients with Engel class I (successful epilepsy surgery) and 48 patients with Engel class II through IV as unsuccessful epilepsy surgery. The demographic data are shown in Tables 1 and 2.

**Table 1.** Categorical Variables.

| *Categorical Variables* | *Number* | *%* |
|---|---|---|
| ***History of febrile seizure*** | | |
| *Yes* | 48 | 27.27% |
| *No* | 128 | 72.73% |
| ***Family history of epilepsy*** | | |
| *Yes* | 33 | 18.75% |
| *No* | 143 | 81.25% |
| ***History of head trauma*** | | |
| *Yes* | 41 | 23.30% |
| *No* | 135 | 76.70% |
| ***Seizure frequency*** | | |
| *Daily* | 56 | 31.82% |
| *Weekly* | 79 | 44.89% |
| *Monthly* | 29 | 16.48% |
| *Yearly* | 8 | 4.55% |
| *Seasonal* | 4 | 2.27% |
| ***Focal to bilateral tonic-clonic seizures*** | | |
| *Yes* | 67 | 38.07% |
| *No* | 109 | 61.93% |
| ***Aura*** | | |
| *No* | 64 | 36.36% |
| *Yes* | 112 | 63.64% |
| ***Lesion location*** | | |
| *Extra-Temporal* | 36 | 20.45% |
| *Temporal* | 140 | 79.55% |
| ***ECoG*** | | |
| *No* | 154 | 87.50% |
| *Yes* | 22 | 12.50% |
| ***MRI findings*** | | |
| *Mesial temporal sclerosis* | 100 | 56.82% |
| *Focal cortical dysplasia* | 19 | 10.80% |
| *Gliosis* | 29 | 16.48% |
| *Tumor* | 22 | 12.50% |
| *Cavernous Angioma* | 6 | 3.41% |
| ***Seizure free*** | | |
| *No* | 48 | 27.27% |
| *Yes* | 128 | 72.73% |

**Table 2.** Numerical Variables.

| ***Numerical Variables*** | ***mean*** | ***std*** | ***min*** | ***max*** |
|---|---|---|---|---|
| *Age at the time of surgery* | 30.54545455 | 9.219122 | 16 | 56 |
| *Age onset of epilepsy* | 13.80208333 | 8.605432 | 0.5 | 45 |
| *seizure duration* | 16.65625 | 9.773712 | 0 | 51 |

std: standard deviation, min: minimum, max: maximum.

## 4. Results

For precise evaluation of the results, the parameters accuracy, precision, recall and, F1-score are calculated using Leave-One-Out method. It should be noted, accuracy shows the performance of the classifiers, but the details of the results should be evaluated with other terms, especially in imbalanced data. For example, if the accuracy of the results is taken into account, support vector machines with a linear kernel have the best results. But classifiers are not accurate in recognizing the minor class. In class zero, the support vector machine with the linear kernel yielded 76.1 percent accuracy, while the recall has a very modest value of 0.23. As the table shows, other classifiers also have faced the same problem. In Figure 1, the numerical parameters: Age and Seizure-Frequency are visualized. As can be seen in both forms, the data are quite complex and intertwined due to the existence of batch variables, we see isolation points of both classes.

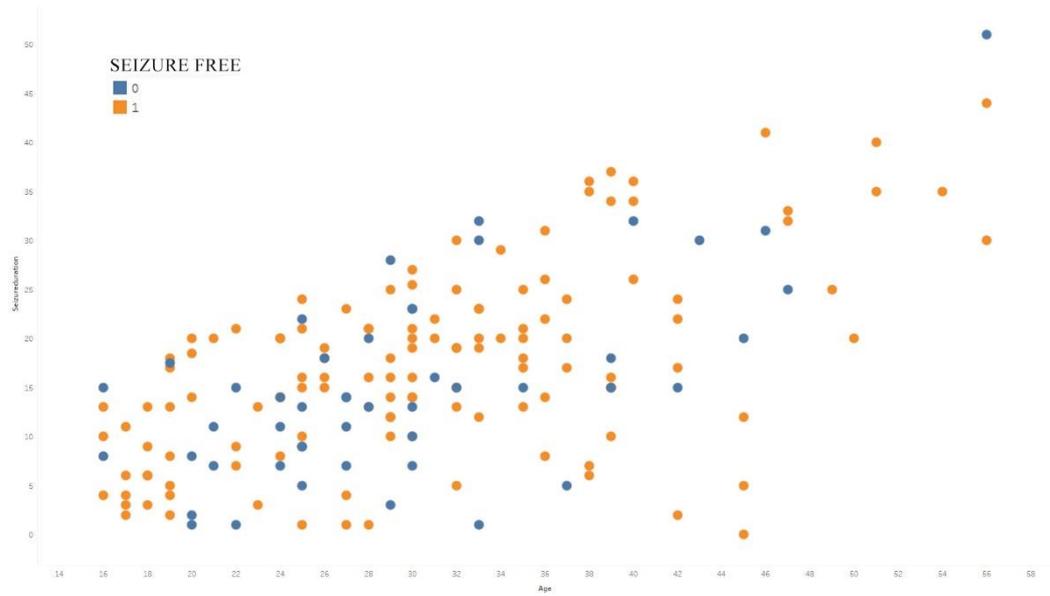

**Figure 1.** Numerical Features.

Table 4 shows the rates of Precision, Recall, and F1-Score after application of over-sampling. We see an improvement in the accuracy of the results as well as the recall and F1-Score index. In methods Decision Tree, Random Forest, Multilayer Perceptron and Gradient Boosting, the results are increased to about 85% and more, as well as SVM with RBF kernel, has a remarkable growth of 96.5% accuracy.

In order to ensure the robustness of the methods we used K-Fold Cross-Validation to measure 5 methods that had over 84% accuracy. This method randomly partitioned data by k subsamples. The k is set to 8 in this experiment. A sample is considered as the test and the remaining is used for the train. Finally, the Average, Standard Deviation, Best, and Worst results are considered for comparison (Table 5). While the SVM (kernel=RBF) has 91.8% mean and 100% best, Multilayer Perceptron has the worst standard deviation and accuracy in terms of mean, best, and worst. On the other hand, the Random Forest with a standard deviation of 4.62 and a small interval between the best and the Worst answer, as well as the mean value accuracy of 86.72%, has stable results.

**Table 3.** Supervised Classifier Results.

| Classifier | Class | Precision | Recall | F1-score | Accuracy |
|---|---|---|---|---|---|
| **Decision Tree** | 0 | 0.33 | 0.33 | 0.33 | 63.1% |
| | 1 | 0.75 | 0.74 | 0.75 | |
| **Random Forest** | 0 | 0.41 | 0.19 | 0.26 | 70.5% |
| | 1 | 0.75 | 0.90 | 0.82 | |
| **Multilayer Perceptron** | 0 | 0.50 | 0.29 | 0.37 | 72.7% |
| | 1 | 0.77 | 0.89 | 0.83 | |
| **Logistic Regression** | 0 | 0.62 | 0.27 | 0.38 | 75.6% |
| | 1 | 0.77 | 0.94 | 0.85 | |
| *K-Nearest Neighbors* | 0 | 0.37 | 0.15 | 0.21 | 69.9% |
| | 1 | 0.74 | 0.91 | 0.81 | |
| *Gradient Boosting* | 0 | 0.41 | 0.38 | 0.39 | 68.2% |
| | 1 | 0.77 | 0.80 | 0.78 | |
| *Support Vector Machine (kernel = linear)* | 0 | 0.69 | 0.23 | 0.34 | 76.1% |
| | 1 | 0.77 | 0.93 | 0.85 | |
| *Support Vector Machine (kernel = Poly)* | 0 | 0.0 | 0.0 | 0.0 | 72.7% |
| | 1 | 0.73 | 1.0 | 0.84 | |
| | 0 | 0.0 | 0.0 | 0.0 | 72.7% |

| | | | | |
|---|---|---|---|---|
| *Support Vector Machine (kernel = RBF)* | 1 | 0.73 | 1.0 | 0.84 | |
| *Support Vector Machine (kernel = Sigmoid)* | 0 | 0.0 | 0.0 | 0.0 | 72.7% |
| | 1 | 0.73 | 1.0 | 0.84 | |

**Table 4.** Supervised Classifier Results after Oversampling.

| Classifier | class | Precision | Recall | F1-score | Accuracy |
|---|---|---|---|---|---|
| **Decision Tree** | 0 | 0.78 | 0.96 | 0.86 | 84.8% |
| | 1 | 0.95 | 0.73 | 0.83 | |
| **Random Forest** | 0 | 0.85 | 0.94 | 0.89 | 88.3% |
| | 1 | 0.93 | 0.83 | 0.88 | |
| **Multilayer Perceptron** | 0 | 0.82 | 0.95 | 0.88 | 87.1% |
| | 1 | 0.94 | 0.80 | 0.86 | |
| **Logistic Regression** | 0 | 0.67 | 0.61 | 0.64 | 65.2% |
| | 1 | 0.64 | 0.70 | 0.67 | |
| *K-Nearest Neighbors* | 0 | 0.62 | 0.70 | 0.66 | 63.7% |
| | 1 | 0.66 | 0.57 | 0.61 | |
| *Gradient Boosting* | 0 | 0.81 | 0.92 | 0.86 | 85.2% |
| | 1 | 0.91 | 0.78 | 0.84 | |
| **Support Vector Machine** (kernel = linear) | 0 | 0.70 | 0.52 | 0.60 | 64.8% |
| | 1 | 0.62 | 0.77 | 0.69 | |
| **Support Vector Machine** (kernel = Poly) | 0 | 0.59 | 0.98 | 0.74 | 65.6% |
| | 1 | 0.95 | 0.33 | 0.49 | |
| **Support Vector Machine** (kernel = RBF) | 0 | 0.99 | 0.94 | 0.96 | 96.5% |
| | 1 | 0.94 | 0.99 | 0.97 | |
| **Support Vector Machine** (kernel = Sigmoid) | 0 | 0.56 | 0.49 | 0.52 | 55.1% |
| | 1 | 0.55 | 0.61 | 0.58 | |

**Table 5.** K-Fold Cross-Validation Results.

| Classifier | Mean | Standard Deviation | Best | Worst |
|---|---|---|---|---|
| **Decision Tree** | 83.20% | 6.04 | 93.75% | 71.88% |
| **Random Forest** | 86.72% | 4.62 | 93.75% | 81.25 |
| **Multilayer Perceptron** | 64.06% | 9.50 | 71.88% | 43.75% |
| *Gradient Boosting* | 82.81% | 7.49 | 96.88% | 71.88% |
| *Support Vector Machine (kernel = RBF)* | 91.8% | 6.98 | 100% | 75.0% |

## 5. Discussion

In our study we proposed the use of a machine learning approach based on supervised classification and feature subset selection data mining to predict the outcome of epilepsy surgery.

Patients with drug-resistant epilepsy who are surgical candidates undergo time-consuming and costly examinations.[10] Although surgery is beneficial in reducing the number and frequency of seizures, the results of surgeries show that a considerable proportion of these patients still have seizures following surgery.[11] As a result, identifying the predictors of surgery outcomes in drug-resistant patients is critical. The data of epileptic patients is complicated, and any links and patterns must be evaluated in precise ways. Machine learning and data mining techniques have recently become popular in medical applications.[12] Machine learning can be used in surgery to predict surgical success or even patient mortality based on disease progression and clinical patient reports.[8,13] The adoption of machine learning modelling in epilepsy surgery was assessed in this study due to the significant potential of machine learning in predicting epilepsy surgery and the lack of a comprehensive review at this time.

We tested the capacity of supervised classifiers to predict the outcome of epilepsy surgery using machine learning tools. Data were examined using various classification algorithms, including Decision Tree, Random Forest, Multilayer Perceptron, Logistic Regression, K-Nearest Neighbors, Gradient Boosting, and Support Vector Machine, as shown in the results. Leave-One-Out and K-fold Cross-Validation methods were used to model the data in this investigation. The models were evaluated using seven approaches, including the Leave-One-Out method, which is ideal for datasets with a little amount of data. According to the kernels linked to support vector machines, a total of ten models were presented. The robustness of the 5 models was assessed using the k-fold cross-validation approach. Our findings are consistent with prior research, which found that SVM with RBF kernel was the most effective classifier in both evaluation models for predicting the success of epilepsy surgery.[14]

It should be noted, accuracy shows the performance of the classifiers, but the details of the results should be evaluated with other terms, especially in imbalanced data. For example, if the accuracy of the results is taken into account, support vector machines with a linear kernel have the best results. But classifiers are not accurate in recognizing the minor class. However, the support vector machine with the linear kernel yielded 76.1% in terms of accuracy, the recall has a very small value with a 0.23 rate in class zero. As the table shows, other classifiers have faced the same problem.

The findings of this study revealed that a machine learning-based presurgical model based on patients' clinical features may accurately predict the outcome of epilepsy surgery in patients with drug-resistant lesional epilepsy. As acknowledged, among the classifiers tested, SVM produced the most accurate results. Our algorithm could accurately predict results in 96.7 percent of TLE patients and 79.5 percent of ETLE cases using ten clinical features. Pre-operative clinical characteristics such as a history of febrile seizure, family history of epilepsy, history of head trauma, seizure frequency, presence of aura, presence of focal to bilateral seizure, lesion location and type, ECoG, age of onset, age at the time of surgery, and epilepsy duration were examined in the current study. Although we did not assess the intensity of each variable, introducing a new feature showing that all of these clinical parameters potentially influence surgery results improved the model's accuracy. A set of three factors, which includes the existence of aura, focal to bilateral seizures, and epilepsy duration, accounted for 85 percent of the model's accuracy. Which can indicate the development of more extensive seizure network in the brain and therefore less responsiveness to focal resection.

Despite the fact that surgery for intractable epilepsy is a helpful treatment option, the rate of epilepsy surgery has not increased in the recent two decades.[15] Small hospitals, general neurologists' unwillingness to recommend patients to epilepsy surgery centres, and a lack of trust among patients are the main causes for this therapy option's underutilization.

Therefore, improved results can help build trust and expand the use of this technique. Many attempts have been undertaken to investigate the predictors of epilepsy surgical results in this way. Several clinical studies have been conducted to describe these predictors, with the results revealing a variety of epilepsy surgical outcome predictors linked to clinical parameters.[16–18] However, these investigations produced contradictory results and were unable to characterise integrated outcome factors across all studies. To establish a set of combined characteristics that can predict surgery outcome, these investigations used statistical models, principally regression. While some of these research found that some characteristics had a positive predictive value,[19–21] others found no effect.[22–24] In this context, building outcome prediction models in epilepsy surgery is a difficulty.

Epilepsy surgery nomograms have been developed as a result of previous efforts. These nomograms were created using statistical models and could predict surgery outcomes in the short term based on clinical characteristics. The Cleveland Clinic's nomogram has shown 85 percent accuracy in predicting surgical success using nine clinical variables: sex, seizure frequency, presence of focal to bilateral seizures, pathology, type of surgery, side of surgery, age-onset, age at the time of surgery, and epilepsy duration.[25,26]

The chances of achieving seizure independence after surgery were predicted to be between 30 and 90 percent.[27,28] In a study, specialists were able to accurately anticipate outcomes in the vast majority (83 percent) of surgical patients before they were offered therapy.[4] Even while nomograms have an accuracy of 85 percent, which is greater than experts' estimates, the difference is not statistically significant.[5]

New challenges in epilepsy surgery outcome prediction have arisen as a result of the need to advance models that can predict surgical results much better than experts.[29] Advances in machine learning models may be able to address these issues. Machine learning models have the advantage of being dynamic, in that the effect of variables is evaluated in conjunction with one another while becoming more accurate as new data is added.

## 6. Conclusions

In lesional epilepsy cases, it appears that using merely the patients' clinical characteristics to predict surgery outcome with high accuracy might be done without the need for more advanced tests. It's worth noting that our model could only predict the short-term outcome of epilepsy surgery; separate models for long-term outcome prediction are needed. Overall, progress in the development of machine learning-based prediction models offers optimism for personalised medicine access.

## 7. Limitations

Spaces are created when a little amount of data is combined with a high number of variables, resulting in isolated points. On the other hand, high-accuracy modelling is problematic due to the near boundaries between classes and the imbalanced data.

## 8. Future directions

Future research should look at not only seizure recurrence but also cognitive changes to predict surgery results. Many research have mostly focused on predicting surgical results in short-term intervals of 1 to 5 years after surgery, indicating a significant need for models that can predict long-term surgical outcomes. Because the rate of successful surgery outcomes in lesional epilepsy is higher than in non-lesional epilepsies, it may be more advantageous to forecast surgery outcomes in this latter group.

**Authors contribution:** Z J: Conceptualization, Investigation, Project administration, Resources, Visualization, Original draft preparation, Review and editing. J M H: Resources, Investigations, Supervision, Review and editing. H M: Investigations, Resources, Review and editing. R B: Investigations, Resources, Review and editing. A R Q: Data curation, Formal analysis, Software, Review and editing. S S G: Validation, Methodology, Review and editing.   A M H: Review and editing. M B: Review and editing. S-A S Z: Supervision, Validation, Project administration, Original draft preparation, Review and editing.

**Ethical publication statement:** We confirm that we have read the Journal's position on issues involved in ethical publication and affirm that this report is consistent with those guidelines.

**Data availability statement:** The data used in this study is available upon request from corresponding. Due to privacy and ethical concerns, the data are not available to the general public.

**Funding statement:** This study has not received any funding.

**Ethics approval statement:** The Regional Bioethics Committee of Isfahan University of Medical Sciences has approved this study (verification code: IR.MUI.MED.REC.1399.669).   **Patient consent statement:** All the patients signed the written informed consent.

**Disclosure of conflict of interest:** None of the authors have any conflict of interest to disclose.

**References**

1. Noachtar S, Borggraefe I. Epilepsy surgery: a critical review. Epilepsy Behav. 2009 May;15(1):66–72.
2. Anyanwu C, Motamedi GK. Diagnosis and Surgical Treatment of Drug-Resistant Epilepsy. Brain Sci. 2018 Mar 21;8(4):E49.
3. Rao MB, Arivazhagan A, Sinha S, Bharath RD, Mahadevan A, Bhat M, et al. Surgery for drug-resistant focal epilepsy. Ann Indian Acad Neurol. 2014 Mar;17(Suppl 1):S124–31.
4. Witt JA, Krutenko T, Gädeke M, Surges R, Elger CE, Helmstaedter C. Accuracy of expert predictions of seizure freedom after epilepsy surgery. Seizure. 2019 Aug 1;70:59–62.
5. Gracia CG, Chagin K, Kattan MW, Ji X, Kattan MG, Crotty L, et al. Predicting seizure freedom after epilepsy surgery, a challenge in clinical practice. Epilepsy Behav. 2019 Jun;95:124–30.
6. Sirven JI. Epilepsy: A Spectrum Disorder. Cold Spring Harb Perspect Med. 2015 Sep;5(9):a022848.
7. Rasheed K, Qayyum A, Qadir J, Sivathamboo S, Kwan P, Kuhlmann L, et al. Machine Learning for Predicting Epileptic Seizures Using EEG Signals: A Review. IEEE Rev Biomed Eng. 2021;14:139–55.
8. Senders JT, Staples PC, Karhade AV, Zaki MM, Gormley WB, Broekman MLD, et al. Machine Learning and Neurosurgical Outcome Prediction: A Systematic Review. World Neurosurg. 2018 Jan;109:476-486.e1.
9. Jason Brownlee. Imbalanced Classification with Python: Better Metrics, Balance Skewed Classes, Cost-Sensitive Learning. 2021.
10. Khoo A, Martin L, Tisi J de, O'Keeffe AG, Sander JW, Duncan JS. Cost of pre-surgical evaluation for epilepsy surgery: A single-center experience. Epilepsy Res. 2022 May;182:106910.
11. Mohan M, Keller S, Nicolson A, Biswas S, Smith D, Farah JO, et al. The long-term outcomes of epilepsy surgery. PLOS ONE. 2018 May 16;13(5):e0196274.
12. Garg A, Mago V. Role of machine learning in medical research: A survey. Computer Science Review. 2021 May 1;40:100370.
13. Elfanagely O, Toyoda Y, Othman S, Mellia JA, Basta M, Liu T, et al. Machine Learning and Surgical Outcomes Prediction: A Systematic Review. Journal of Surgical Research. 2021 Aug 1;264:346–61.
14. Memarian N, Kim S, Dewar S, Engel J, Staba RJ. Multimodal data and machine learning for surgery outcome prediction in complicated cases of mesial temporal lobe epilepsy. Comput Biol Med. 2015 Sep;64:67–78.
15. Engel J. The current place of epilepsy surgery. Curr Opin Neurol. 2018 Apr;31(2):192–7.
16. Tonini C, Beghi E, Berg AT, Bogliun G, Giordano L, Newton RW, et al. Predictors of epilepsy surgery outcome: a meta-analysis. Epilepsy Res. 2004 Nov;62(1):75–87.
17. Sarkis RA, Jehi L, Bingaman W, Najm IM. Seizure worsening and its predictors after epilepsy surgery. Epilepsia. 2012;53(10):1731–8.
18. Englot DJ, Chang EF. Rates and predictors of seizure freedom in resective epilepsy surgery: an update. Neurosurg Rev. 2014 Jul 1;37(3):389–405.
19. Wiebe S, Blume WT, Girvin JP, Eliasziw M. A Randomized, Controlled Trial of Surgery for Temporal-Lobe Epilepsy. New England Journal of Medicine. 2001 Aug 2;345(5):311–8.
20. Lamberink HJ, Otte WM, Blümcke I, Braun KPJ, Aichholzer M, Amorim I, et al. Seizure outcome and use of antiepileptic drugs after epilepsy surgery according to histopathological diagnosis: a retrospective multicentre cohort study. The Lancet Neurology. 2020 Sep 1;19(9):748–57.
21. Englot DJ, Wang DD, Rolston JD, Shih TT, Chang EF. Rates and predictors of long-term seizure freedom after frontal lobe epilepsy surgery: a systematic review and meta-analysis: Clinical article. Journal of Neurosurgery. 2012 May 1;116(5):1042–8.

22. O'Dwyer R, Byrne R, Lynn F, Nazari P, Stoub T, Smith MC, et al. Age is but a number when considering epilepsy surgery in older adults. Epilepsy Behav. 2019 Feb;91:9–12.

23. Spencer SS, Berg AT, Vickrey BG, Sperling MR, Bazil CW, Shinnar S, et al. Predicting long-term seizure outcome after resective epilepsy surgery: The Multicenter Study. Neurology. 2005 Sep 27;65(6):912–8.

24. Hitti FL, Piazza M, Sinha S, Kvint S, Hudgins E, Baltuch G, et al. Surgical Outcomes in Post-Traumatic Epilepsy: A Single Institutional Experience. Oper Neurosurg (Hagerstown). 2020 Jan 1;18(1):12–8.

25. Garcia Gracia C, Yardi R, Kattan MW, Nair D, Gupta A, Najm I, et al. Seizure freedom score: A new simple method to predict success of epilepsy surgery. Epilepsia. 2015;56(3):359–65.

26. Jehi L, Yardi R, Chagin K, Tassi L, Russo GL, Worrell G, et al. Development and validation of nomograms to provide individualised predictions of seizure outcomes after epilepsy surgery: a retrospective analysis. The Lancet Neurology. 2015 Mar 1;14(3):283–90.

27. Elsharkawy AE, Pannek H, Schulz R, Hoppe M, Pahs G, Gyimesi C, et al. OUTCOME OF EXTRATEMPORAL EPILEPSY SURGERY EXPERIENCE OF A SINGLE CENTER. Neurosurgery. 2008 Sep 1;63(3):516–26.

28. Feis DL, Schoene-Bake JC, Elger C, Wagner J, Tittgemeyer M, Weber B. Prediction of post-surgical seizure outcome in left mesial temporal lobe epilepsy. NeuroImage: Clinical. 2013 Jan 1;2:903–11.

29. Fassin AK, Knake S, Strzelczyk A, Josephson CB, Reif PS, Haag A, et al. Predicting outcome of epilepsy surgery in clinical practice: Prediction models vs. clinical acumen. Seizure. 2020 Mar 1;76:79–83.

30. Sadegh-Zadeh, S. A. (2019). Computational methods toward early detection of neuronal deterioration (Doctoral dissertation, University of Hull).

31. Sadegh-Zadeh, Seyed-Ali, Chandrasekhar Kambhampati, and Darryl N. Davis. "Ionic Imbalances and Coupling in Synchronization of Responses in Neurons." J 2.1 (2019): 17-40.

32. Sadegh-Zadeh, Seyed-Ali, and Chandrasekhar Kambhampati. "Computational Investigation of Amyloid Peptide Channels in Alzheimer's Disease." J 2.1 (2018): 1-14.

33. Widjaja, Elysa, et al. "Seizure outcome of pediatric epilepsy surgery: systematic review and meta-analyses." Neurology 94.7 (2020): 311-321.